\def\year{2021}\relax
\documentclass[letterpaper]{article} 
\usepackage{aaai21}  
\usepackage{times}  
\usepackage{helvet} 
\usepackage{courier}  
\usepackage[hyphens]{url}  
\usepackage{graphicx} 
\usepackage{graphicx}  
\usepackage{natbib}  
\usepackage{caption} 
\usepackage{bm}
\usepackage{xcolor}
\usepackage{multirow}
\usepackage{makecell}
\usepackage{textcomp}
\usepackage{amssymb}
\usepackage{amsmath}

\title{Structured Co-reference Graph Attention for Video-grounded Dialogue}

\author{
	Junyeong Kim, Sunjae Yoon, Dahyun Kim, Chang D. Yoo\\
}
\affiliations{
    Korea Advanced Institute of Science and Technology (KAIST) \\
    \{junyeong.kim, sunjae.yoon, dahyun.kim, cd\_yoo\}@kaist.ac.kr
}

\begin{document}

\maketitle

\begin{abstract}
A video-grounded dialogue system referred to as the Structured Co-reference Graph Attention (SCGA) is presented for decoding the answer sequence to a question regarding a given video while keeping track of the dialogue context.
Although recent efforts have made great strides in improving the quality of the response, performance is still far from satisfactory. The two main challenging issues are as follows: (1) how to deduce co-reference among multiple modalities and (2) how to reason on the rich underlying semantic structure of video with complex spatial and temporal dynamics.
To this end, SCGA is based on (1) Structured Co-reference Resolver that performs dereferencing via building a structured graph over multiple modalities, (2) Spatio-temporal Video Reasoner that captures local-to-global dynamics of video via gradually neighboring graph attention.
SCGA makes use of pointer network to dynamically replicate parts of the question for decoding the answer sequence.
The validity of the proposed SCGA is demonstrated on AVSD@DSTC7 and AVSD@DSTC8 datasets, a challenging video-grounded dialogue benchmarks, and TVQA dataset, a large-scale videoQA benchmark.
Our empirical results show that SCGA outperforms other state-of-the-art dialogue systems on both benchmarks, while extensive ablation study and qualitative analysis reveal performance gain and improved interpretability.
\end{abstract}

\section{Introduction}

Understanding visual information along with the natural language appears to be a desiderata in our community.
Thus far, notable progress has been made towards bridging the fields of computer vision and natural language processing that includes video moment retrieval \cite{Ma_2020_ECCV}, image-grounded question answering \cite{Antol_2015_ICCV,Anderson_2018_CVPR} / dialogue \cite{Das_2017_CVPR,Vries_2017_CVPR}, and video-grounded question answering \cite{Tapaswi_2016_CVPR,Lei_2018_EMNLP} / dialogue \cite{Alamri_2019_CVPR}.
Among those, we focus on video-grounded dialogue system (VGDS) that allows an AI agent to `observe' (i.e., understand a video) and `converse' (i.e., communicate the understanding in a dialogue).
To be specific, given a video, dialogue history consisting of a series of QA pairs, and a follow-up question about the video, the goal is to infer a free-form natural language answer to the question.
Video-grounded dialogue appears often in many real-world human-computer conversations, and VGDS can potentially provide assistance to various subsections of the population especially to those subgroups suffering from sensory impairments. 
Although recent years has witnessed impressive advancement in performance, current VGDSs are still struggling with the following two challenging issues: (1) how to fully co-reference among multiple modalities, and (2) how to reason on the rich underlying semantic structure of video with complex spatial and temporal dynamics.

\begin{figure}[t]
	\centering
	\includegraphics[width=\linewidth]{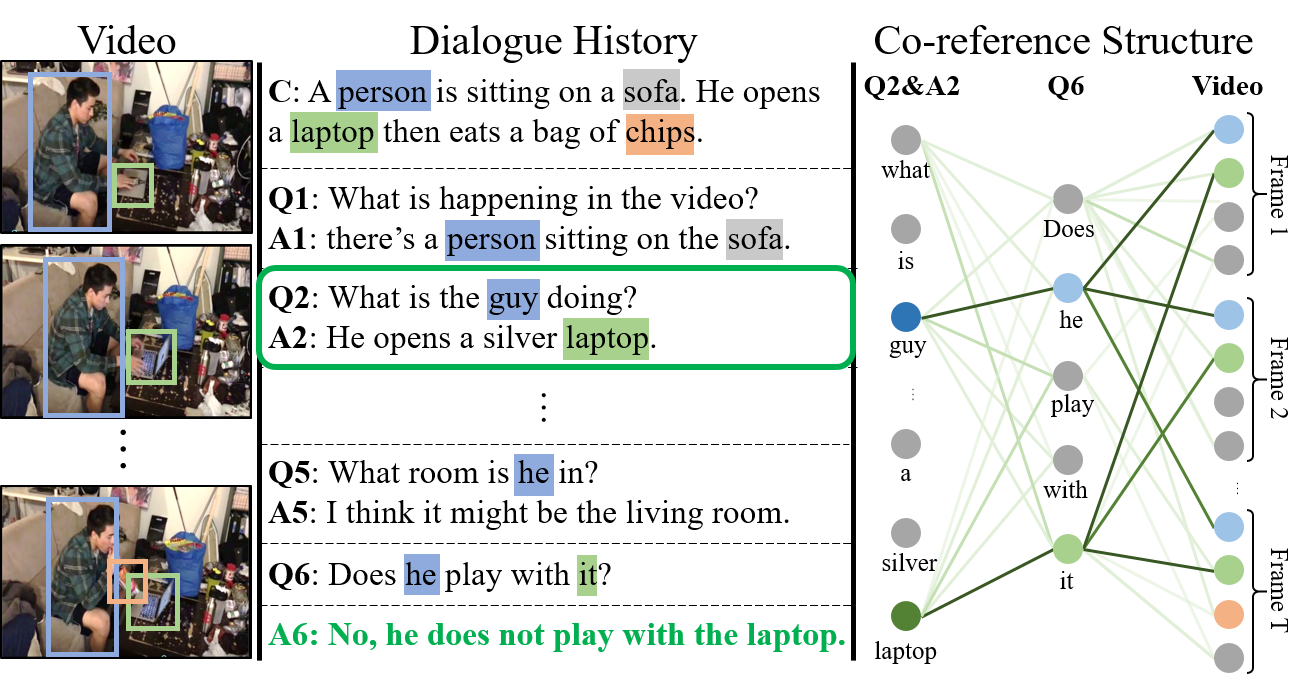}
	\caption{Illustration of the intuition behind SCGA as a video-grounded dialogue system. Left: video with detected objects, Middle: dialogue history, current Q\&A. Most informative history is indicated by green box. Right: structured co-reference graph representing the underlying semantic dependencies between nodes (darker links indicate higher dependencies).} 
	\label{fig:intro}
\end{figure}

%
%
%

The first challenging issue of co-referencing among multiple modalities is illustrated in Figure \ref{fig:intro}: To the question in Q6 ``Does \underline{he} play with \underline{it}?'' contains pronouns for which the noun referent or antecedent must be identified. Henceforth, this task of identifying the pronoun's antecedent will be refer to as "dereferencing". 
Our study shows that nearly all dialogues and 61\% of questions in the audio visual scene-aware dialogue (AVSD) dataset contains at least one pronoun (e.g., ``it'', ``they'') which makes ``dereferencing" indispensable.
Existing VGDSs have treated dialogue history just as any another input modality, but by recognizing and resolving its unique issue regarding pronoun reference, the quality of the output response can be enhanced significantly. For the first time, this paper proposes a VGDS that performs textual and visual co-reference via structured co-reference graph that identifies the pronoun's antecedents with nouns in the prior dialog and also with detected objects in video. 
This approach can be considered as an extension of prior efforts on visual dialogue (VisDial) to resolve visual co-reference issues via attention memory \cite{Seo_2017_NIPS}, reference pool \cite{Kottur_2018_ECCV} and recursive attention \cite{Niu_2019_CVPR}.
%

The second challenging issue is to perform reasoning on rich underlying semantic structure of video with complex spatial and temporal dynamics.
Majority of prior efforts on video-grounded dialogue task utilize holistic video feature from I3D model \cite{Carreira_2017_CVPR} that has been pre-trained on action recognition dataset, and they rely on a fully-connected transformer architecture to implicitly learn to infer the answer \cite{Le_2019_ACL,Le_2020_DSTC8,Lee_2020_DSTC8}.
These efforts underestimate the value of fine-grained visual representation from detector \cite{Ren_2015_NIPS,Vu_2019_NIPS}, consequently lacking the capability to apprehend relevant objects and their relationships and temporal evolution.
%
%
We design a spatio-temporal video graph over object-level representation and perform graph attention for comprehensive understanding of the video.
%

In this paper, we address the aforementioned challenging issues with our Structured Co-reference Graph Attention (SCGA) which is composed of (1) structured co-reference resolver that performs dereferencing via building a structured co-reference graph, (2) spatio-temporal video reasoner that captures local-to-global dynamics of video via gradually neighboring graph attention (GN-GAT).
We first \textit{select} a key dialog history that can resolve the pronoun's antecedent in the follow-up question.
Gumbel-Softmax \cite{Jang_2016_arxiv,Maddison_2016_arxiv} enables us to perform \textit{discrete} attention over dialogue history.
We propose a \textit{bipartite} structured co-reference graph over multiple modalities and perform graph attention to integrate informative semantics from the key dialog history to question and video, as shown in Figure \ref{fig:intro}.
We then build spatio-temporal video graph that represents spatial and temporal relations among objects.
Motivated by recent studies that each head in self-attention independently looks at same global context, learning redundant features \cite{Voita_2019_ACL,Kant_2020_ECCV}, we propose \textit{gradually neighboring} graph attention (GN-GAT) that is guided by constructed video graph.
%
%
Rather than repeatedly calculating self-attention over the common neighborhood, each head looks at a different neighborhood defined by its unique adjacency matrix. Each adjacency matrix will have a unique connectivity that link nodes reached within a fix number of hops. Thus, heads associated with smaller number of hops will consider local context while heads associated a larger number of hops will be looking at the global context. 
%
%
Finally based on the observation that words used in response come from words used in the question (e.g., response to question ``What did he do after closing the window?'' can be ``He [context verb] after closing the window''), a pointer network is incorporated into the response sequence decoder to either decode from a fixed vocabulary set or from words used in the question.

%
%

\section{Related Work}

\subsection{Video-grounded  Dialogues}
Visual Question Answering (VQA) \cite{Antol_2015_ICCV} has been considered as a proxy task to evaluate the model's understanding on vision and language.
In recent years, video-grounded dialogue systems \cite{Alamri_2019_CVPR,Hori_2019_ICASSP} have been proposed to advance VQA to hold meaningful dialogue with humans, grounded on video.
%
%
VGDS incorporating recurrent neural network to encode dialog history is considered in \cite{Hori_2019_ICASSP,Nguyen_2019_DSTC7,Le_2019_DSTC7,Sanabria_2019_DSTC7}.
%
%
Transformer based VGDS has been considered with query-aware attention \cite{Le_2019_ACL}, word-embedding attention decoder \cite{Lee_2020_DSTC8}, and pointer-augmented decoder \cite{Le_2020_DSTC8}.
%
%
VGDS that generates scene-graph every frame and aggregates it over the temporal axis to model fine-grained information flow in videos has also been considered \cite{Geng_2020_DSTC8}. 
These systems would perform better with (1)"dereferencing" capability  and (2) capability to capture and reason on complex spatial and temporal dynamics of the video.
\begin{figure*}[t]
	\centering
	\includegraphics[width=\textwidth]{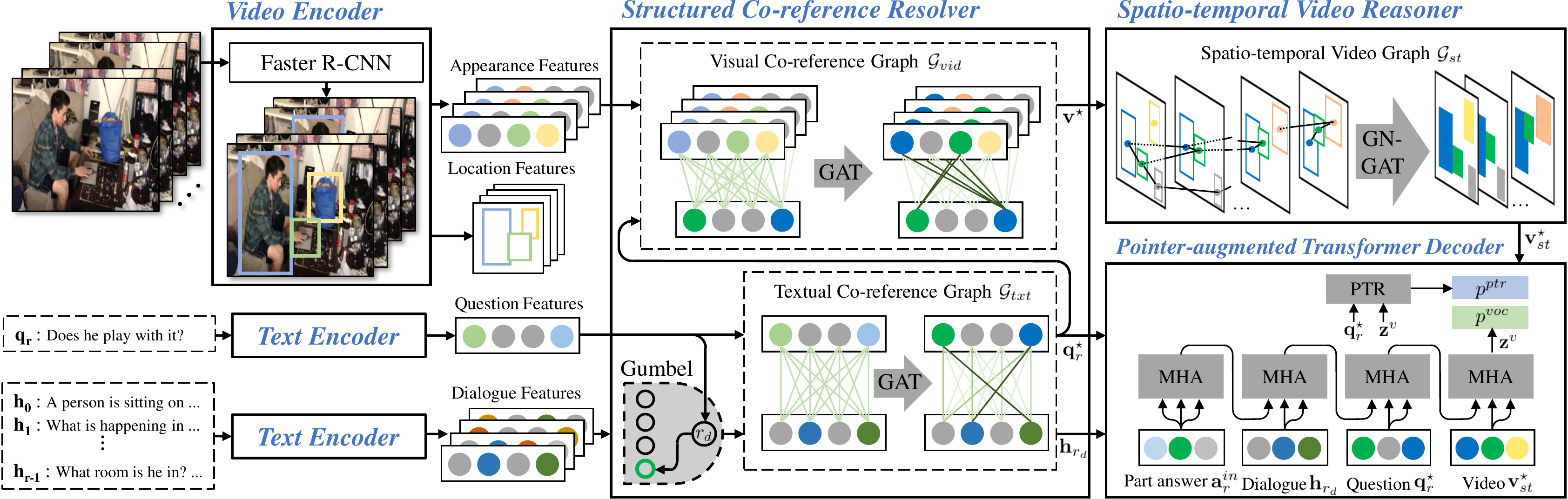}
	\caption{Illustration of Structured Co-reference Graph Attention (SCGA) which is composed of: (1) Input Encoder, (2) Structured Co-reference Resolver, (3) Spatio-temporal Video Reasoner, (4) Pointer-augmented Transformer Decoder.}
	\label{fig:method}
\end{figure*}
\subsection{Co-reference Resolution}
%

Co-reference resolution is a task that was first defined in the linguistic community \cite{Bertsma_2006_ACL}, whose objective is to build association between named entities and references. It would include the task of identifying or associating the pronoun's antecedent from the dialog history and detected objects.
While none of the past VGDSs have explicitly considered co-reference resolution, a number of systems in the VisDial have attempted to resolve visual co-reference.
In \cite{Seo_2017_NIPS}, attention memory stores a sequence of
previous (attention, key) pairs and the most relevant previous pair for the current question is retrieved. 
In \cite{Kottur_2018_ECCV}, a neural module network refers back to entities from previous rounds of dialogue and reuses its associated entities.
In \cite{Niu_2019_CVPR}, the dialog history is browses until the agent has sufficient confidence in the visual co-reference resolution, and refines the visual attention recursively.
%
Taking co-reference resolution to another level from just focusing on visual co-reference, SCGA conducts both textual and visual co-reference via structured co-reference graph.
\subsection{Graph-based Visual Reasoning}
Prior works have used GCN \cite{Kipf_2017_ICLR} or GAT \cite{velickovic_2017_arxiv} to enable relational reasoning for image captioning \cite{Yao_2018_ECCV}, VQA \cite{Teney_2017_CVPR,Li_2019_ICCV}, and VideoQA \cite{Jiang_2020_AAAI}.
Fully-connected graph between objects \cite{Teney_2017_CVPR}, objects / words \cite{Jiang_2020_AAAI} or spatial / semantic graph \cite{Yao_2018_ECCV,Li_2019_ICCV} is constructed to link different objects in relationship.
By contrast, we construct \textit{bipartite} graph between multiple modalities to form structured co-reference graph.
Further, we build video graph not only in spatial-axis but also in temporal axis.
We also provide different role for each head in gradually neighborhood graph attention to prevent learning redundant features.
\section{Method}
Here a formal definition of the video-grounded dialogue task is provided \cite{Alamri_2019_CVPR}.
We are given tuples of $(v, h, q_r)$, consisting of a video $v$, the dialogue history $h = \{c, (q_1, a_1), \cdots, (q_{r-1}, a_{r-1})\}$, and a question $q_r$ asked at current round $r \in \{1, \cdots, R\}$.
The dialogue history itself is a set of question-answer pairs of previous rounds with a caption $c$ in the beginning.
The goal of video-grounded dialogue is to generate free-form natural language answer $a_r$ to the question.
Figure \ref{fig:method} shows a schematic of Structured Co-reference Graph Attention (SCGA), consisting of a Input Encoder, Structured Co-reference Resolver, Spatio-temporal Video Reasoner, and Pointer-augmented Transformer Decoder.
For the Video Encoder, Faster R-CNN \cite{Ren_2015_NIPS} is used to extract sets of objects $v^t = \{v_o^t\}_{o=1}^O$ for each frame of $t \in \{1, \cdots, T\}$, where each object $v_o^t$ is represented with an appearance feature vector $\mathbf{v}_o^t \in \mathbb{R}^{d_v}$ and location feature $\mathbf{b}_o^t \in \mathbb{R}^{d_b}$ ($T=15$, $O=6$, $d_v=2048$, and $d_b=4$) in our experiment.
Each location feature $\mathbf{b}_o^t = [x, y, w, h]$ represents a spatial coordinate, where $[x,y]$ denotes the relative coordinate of top-left point of the b-box while $[w, h]$ denotes the width and height of the box.
For the Text Encoder, we use a trainable token-level embedding layer to map sequence of token indices into $d$-dimensional feature representations.
To incorporate ordering information of source tokens, we apply positional encoding \cite{Vaswani_2017_NIPS} with layer normalization \cite{Ba_2016_arxiv} on top of embedding layer.
The encoded question ($q_r$) and each of dialogue history ($\{h_i\}_{i=1}^{r-1}$) are defined as:
\begin{align}
\mathbf{q}_{r} &= \mbox{LN}(\phi(q_r) + \mbox{PE}(q_r)) \in \mathbb{R}^{N_{q_r} \times d}, \\
\mathbf{h}_i &= \mbox{LN}(\phi(h_i) + \mbox{PE}(h_i)) \in \mathbb{R}^{N_{h_i} \times d},
\end{align}
where $N_x$ denotes the number of tokens of sequence $x$.
The followings sub-sections will explain the details of remaining model components.

\subsection{Structured Co-reference Resolver}
\subsubsection{Textual Co-reference Resolution.}
We observed that there exists one key dialogue history that can resolve co-reference in the current question.
To inject semantic information from the dialogue history into the question token representation for textual co-reference resolution, we first determine a key dialogue history, and then we let question tokens to attend to key dialogue history tokens.
In our framework, we implement those functions via Gumbel-Softmax \cite{Jang_2016_arxiv,Maddison_2016_arxiv} to perform \textit{discrete} attention over dialogue histories and graph attention over \textit{bipartite} graph that connects all of the question tokens to all of the dialogue tokens.
In this manner, question tokens learn to implicitly integrate informative semantics from the key dialogue history to its representation.
%

One can easily suppose that the current question simply follows from the latest dialogue history.
However, sometimes the question requires looking back at an earlier dialogue, which means there are no relationships between the current question and recent dialogue histories.
Inspired by Gumbel-Max trick with continuous softmax relaxations \cite{Niu_2019_CVPR}, we select a most relevant dialogue history $h_{r_d}$ for current question $q_r$.
Our approach is end-to-end trainable while making \textit{discrete} decision, thanks to Gumbel-Softmax.
We first calculate matching score $s_{r,i}$ between question feature $\textbf{q}_r$ and each of dialogue history features $\{\mathbf{h}_0, \cdots, \mathbf{h}_{r-1}\}$:
\begin{align}
e_{r, i} &= f_e([f_q(\mathcal{A}(\mathbf{q}_r)) || f_h(\mathcal{A}(\mathbf{h}_{i}))]), \\
s_{r, i} &= f_s([e_{r_i} || \Delta_{r,i}]),
\end{align}
where $[\cdot || \cdot]$ denotes concatenation operation, $\mathcal{A}$ represents average operation on word-axis, and $f_x$ is a fully-connected layer with input $x$. Here, $\Delta_{r,i} = r-i$ provides distance information between $q_r$ and $h_i$ in the dialogue history. Gumbel-Softmax produces a $r$-dimensional one-hot vector $\mathbf{g}_r$ for \textit{discrete} attention over dialogue histories:
\begin{align}
\mathbf{g}_r &= \mbox{Gumbel\_Softmax}(\mathbf{s}_r), \\
\mathbf{h}_{r_d} &= \sum_{i=0}^{r-1} g_{r,i} \cdot \mathbf{h}_i.
\end{align}
%

We formally define the textual co-reference graph $\mathcal{G}_{txt} = (\mathcal{V}_{txt}, \mathcal{E}_{txt})$ by treating each token from question $q_r$ and related dialogue history $h_{r_d}$ as graph nodes.
In designing a co-reference resolver, we construct a \textit{bipartite} graph and perform graph attention \cite{velickovic_2017_arxiv} to inject useful semantic information from the related dialogue history into query representation.
We first concatenate the $\mathbf{q}_r$ and $\mathbf{h}_{r_d}$ along word-axis to make a heterogeneous node matrix:
\begin{align}
\bm{\mathcal{V}}_{txt} &= [\mathbf{q}_r\ ||\ \mathbf{h}_{r_d}] \in \mathbb{R}^{N_{txt} \times d},
\end{align}
where $N_{txt} = N_{q_r} + N_{h_{r_d}}$ is the number of nodes.
Multi-head self-attention is then performed to model relations between each node and its neighboring nodes. 
For each head $k$, attention coefficient $\alpha_{i,j}^{k}$ denoting the relevance between two linked node $\bm{\mathcal{V}}_i$ and $\bm{\mathcal{V}}_j$ is calculated as:
\begin{align}
\alpha_{i,j}^{k} &= \frac{\exp(\sigma(a^\top_k [W^k \bm{\mathcal{V}}_i || W^k \bm{\mathcal{V}}_j])}{\sum_{n \in \mathcal{N}_i} \exp(\sigma(a^\top_k [W^k \bm{\mathcal{V}}_i || W^k \bm{\mathcal{V}}_n]))}, \label{eq:8}
\end{align}
where $\sigma$($\cdot$) is a nonlinear function such as LeakyReLU, $a_k \in \mathbb{R}^{2d}$ is the attention weight vector, $W^k \in \mathbb{R}^{d \times d}$ is the shared projection matrix, and $\mathcal{N}_i$ is the neighborhood of node $i$.
Graph node features are updated by going through $K$ independent attention mechanisms, and concatenating their output features:
\begin{align}
\bm{\mathcal{V}}_i^\star &= ||_{k=1}^{K} \sigma(\sum_{j \in \mathcal{N}_i} \alpha_{i, j}^k W^k \bm{\mathcal{V}}_j). \label{eq:9}
\end{align}
At the end, we add original $\bm{\mathcal{V}}_i$ to updated $\bm{\mathcal{V}}_i^\star$ and pick nodes corresponding to question tokens to serve as the final co-reference resolved question representation:
\begin{align}
\mathbf{q}_r^\star &= (\bm{\mathcal{V}}_{txt}^\star + \bm{\mathcal{V}}_{txt})[:N_{q_r}] \in \mathbb{R}^{N_{q_r} \times d},
\end{align}
where $[:i]$ denotes slicing operation along node-axis.
\subsubsection{Visual Co-reference Resolution.}
Pipeline of visual co-reference resolution is analogous to textual co-reference resolution.
Video objects learn to implicitly integrate informative semantics from a co-reference resolved question to its representation.
We first project each appearance feature $\mathbf{v}_k^t$ into $d$-dimensional space with linear transformation (where $d$ is the same as in the text embedding).
We also add layer normalization \cite{Ba_2016_arxiv} on top of linear transform to ensure that the appearance feature has same scale as text representation $\mathbf{v}_o^t := \mbox{LN}(W\mathbf{v}_o^t)$.
Again, we construct \textit{bipartite} visual co-reference graph $\mathcal{G}_{vid} = (\mathcal{V}_{vid}, \mathcal{E}_{vid})$ by treating every objects from video and token from co-reference resolved question as graph nodes:
\begin{align}
\bm{\mathcal{V}}_{vid} &= [\mathbf{v}\ ||\ \mathbf{q}_r^\star] \in \mathbb{R}^{N_{vid} \times d},
\end{align}
where $N_{vid} = N_v + N_{q_r}$, and $N_v = T \times O$.
We perform graph attention over graph $\mathcal{G}_{vid}$ and obtain updated graph node $\bm{\mathcal{V}}_{vid}^\star$.
Finally, we pick nodes corresponding to objects to serve as the final co-reference resolved video representation:
\begin{align}
\mathbf{v}^\star &= (\bm{\mathcal{V}}_{vid}^\star + \bm{\mathcal{V}}_{vid})[:N_v] \in \mathbb{R}^{N_v} \times d.
\end{align}

\subsection{Spatio-temporal Video Reasoner}
We first construct a spatio-temporal video graph $\mathcal{G}_{st} = (\mathcal{V}_{st}, \mathcal{E}_{st})$ that represents spatial and temporal among between detected objects, and perform our proposed \textit{gradually neighboring} graph attention (GN-GAT) to reason on rich underlying semantic structure of video with complex spatial and temporal dynamics.
Recent studies \cite{Voita_2019_ACL,Kant_2020_ECCV} show that multi-head self attention bares a limitation in learning redundant features due to repeated usage of same input context for every attention heads.
Rather than repeatedly calculating self-attention over same neighborhood, we propose GN-GAT that each head considers different adjacency matrix whose connectivity gradually increases with distance with respect to the graph nodes.
We can effectively model and reason on local-to-global context of spatial and temporal dynamics of video.
%

While using co-reference resolved video representation $\mathbf{v}^\star \in \mathbb{R}^{N_v \times d}$ as graph node $\mathcal{V}_{st}$, we define two sets of edge matrices, $\{E_{t}\}_{t=1}^T$ that capture spatial relations within each frame and $\{E_{t}^{t+1}\}_{t=1}^{T-1}$ that capture temporal relations between adjacent frames, to build graph edge $\mathcal{E}_{st}$. 
Here, we use location feature $\mathbf{b}_{o}^{t}$ to obtain $E_{t}$ and $E_{t}^{t+1}$.
Criterion for spatial relation matrix $E_t \in \mathbb{R}^{O \times O}$ is defined as $\max(\Delta_x, \Delta_y) < \tau_s$.
More concretely, $E_t[i, j] = 1$ if $i$-th object and $j$-th object in frame $t$ are close enough to match above criterion.
Criterion for temporal relation matrix $E_t^{t+1}$ is defined as $\max(\Delta_x, \Delta_y) < \tau_t$ with same object label.
Again, $E_t^{t+1}[i, j] = 1$ if $i$-th object in frame $t$ and $j$-th object in frame $t+1$ are close enough and have same object label.
Finally, graph edge $\mathcal{E}_{st} \in \mathbb{R}^{N_v \times N_v}$ is constructed as follows:

\begin{align}
\mathcal{E}_{st} &= \begin{bmatrix} 
E_1         & E_1^2       & \mathbf{0} & \mathbf{0} & \cdots  & \mathbf{0} \\
E_1^{2\top} & E_2         & E_2^3      & \mathbf{0} & \cdots  & \mathbf{0} \\
\mathbf{0}  & E_2^{3\top} & E_3        & E_3^4      & \cdots  & \mathbf{0} \\
\vdots      &             &            & \ddots     &         &            \\
\mathbf{0}  & \cdots      & \mathbf{0} & \mathbf{0} & E_{T-1}^{T\top} & E_T        \\
\end{bmatrix}.
\end{align}

The GN-GAT performs reasoning on constructed spatio-temporal video graph $\mathcal{G}_{st}$.
Adjacency matrix for distance $n$ among graph nodes can be calculated as boolean of $n$-th power of $\mathcal{E}_{st}$: 
\begin{align}
A_n &= \mbox{Bool}(\mathcal{E}_{st}^n).
\end{align}
Each head of GN-GAT looks at adjacency matrix $A_n$ with different distance $n$.
In this way, we can effectively learn from local (i.e., lower distance $n$) to global (i.e., higher distance $n$) context of video gradually within a single graph attention layer.
We assign more heads to higher $n$ since global context consists much denser neighborhood compared to local context.
Finally, we obtain $\mathbf{v}^\star_{st}$ by performing GN-GAT:
\begin{align}
\mathbf{v}^\star_{st} &= \mbox{GN-GAT}(\mathbf{v}^\star) \in \mathbb{R}^{N_v \times d},
\end{align}
where $\mbox{GN-GAT}$ is formulated as Equation \ref{eq:8},\ref{eq:9} with different neighborhood defined by adjacency matrix for each head.

\subsection{Pointer-augmented Transformer Decoder}
Answer response is decoded by incorporating the question and video representations from the preceding model components.
Following prior work \cite{Hori_2019_ICASSP}, we decode answer tokens in autoregressive manner.
We design a Transformer decoder consists of 4 attention layers: masked self-attention to partially generated answer so far ($\mathbf{a}_r^{in}$), guided-attention to selected history ($\mathbf{h}_{r_d}$), guided-attention to question ($\mathbf{q}_r^\star$) from structured co-reference resolver, and guided attention to video ($\mathbf{v}^{\star}_{st}$) from spatio-temporal video reasoner.
We further augment decoder with dynamic pointer network \cite{Vinyals_2015_NIPS,Hu_2020_CVPR} to either decode token from fixed vocabulary or copy from question words based on the intuition that question tokens can form a structure of answer (e.g., ``He [context verb] after closing the window'' for question ``What does he do after closing the window?'').

Each attention on transformer decoder is multi-head attention \cite{Vaswani_2017_NIPS} on query, key, and value tensors:
$\mbox{Attention}(Q, K, V)$. 
%
%
Our transformer decoder can be formulated as:
\begin{align}
\mathbf{z}^{a} &= \mbox{Attention}(\mathbf{a}_{r}^{in}, \mathbf{a}_{r}^{in}, \mathbf{a}_{r}^{in}), \\
\mathbf{z}^{h} &= \mbox{Attention}(\mathbf{z}^{a}, \mathbf{h}_{r_d}, \mathbf{h}_{r_d}), \\
\mathbf{z}^{q} &= \mbox{Attention}(\mathbf{z}^{h}, \mathbf{q}_r^\star, \mathbf{q}_r^\star), \\
\mathbf{z}^{v} &= \mbox{Attention}(\mathbf{z}^{q}, \mathbf{v}^{\star}_{st}, \mathbf{v}^{\star}_{st}),
\end{align}
where $\mathbf{a}_{r}^{in} \in \mathbb{R}^{j \times d}$ is partially generated answer at $j$-th decoding step embedded with text encoder. 
Note that we mask the first self-attention layer to ensure causality in answer decoding.
At $j$-th decoding step, we either choose word index from fixed vocabulary distribution $p_{j}^{voc}$ or dynamic pointer distribution $p_{j}^{ptr}$ through argmax operation on $p_{j} = [p_j^{voc} || p_j^{ptr}]$:
%
\begin{align}
p_{j}^{voc} &= g^{voc}(\mathbf{z}^v_j) \in \mathbb{R}^{||V||},\\
p_{j}^{ptr} &= g^{ptr}_q(\mathbf{q}_r^\star)^\top g^{ptr}_z(\mathbf{z}^v_j) \in \mathbb{R}^{N_{q_r}},
\end{align}
where $g^{voc}$ is a linear layer to vocabulary size $||V||$-dimension, and $g^{ptr}_x$ is a linear layer to $d$-dimension.
Logits from dynamic pointer network is obtained through bilinear interaction between question token representation (i.e., $g^{ptr}_q(\mathbf{q}_r^\star)$) and decoder output (i.e., $g^{ptr}_z(\mathbf{z}^v_j)$).

\subsection{Optimization}
During training, we use teacher-forcing \cite{Lamb_2016_NIPS} to supervise each decoding steps, i.e., ground-truth tokens are used as decoder input.
We train the model with multi-label binary cross entropy loss over concatenated token distribution $p_j$, since answer token can appear on both fixed vocabulary and question tokens. 
We add two special tokens to our fixed answer vocabulary, \texttt{<bos>} and \texttt{<eos>}, where \texttt{<bos>} is used as first step of decoder to indicates the beginning of sentence and \texttt{<eos>} denotes the end of sentence to stop the decoding process.

\section{Experiments}

\subsection{Datasets}
We validate our proposed SCGA on two recent datasets.
\subsubsection{AVSD \cite{Alamri_2019_CVPR}} 
is a widely used benchmark dataset for video-grounded dialogue, which are collected on the Charades \cite{Sigurdsson_2016_ECCV} human-activity dataset.
It contains 7,659, 1,787, 1,710 dialogues for training, validation and test, respectively.
Each dialogue contains 10 dialogue turns, and each turn consists of a question and target response.
For evaluation, 6 reference responses are provided.
We provide experimental results on both AVSD@DSTC7 \cite{Alamri_2019_DSTC7} and AVSD@DSTC8 \cite{Hori_2020_DSTC8} challenge benchmark.

\subsubsection{TVQA \cite{Lei_2018_EMNLP}}
is a large-scale benchmark dataset for multi-modal video question answering, which consists multiple-choice QA pairs for short video clips and corresponding subtitles.
It contains 122,039, 15,252, 7,623 QAs for training, validation and test, respectively.
To fit our problem setting, we made some modifications to TVQA.
Among multiple answer candidates, we select correct one to be a target response.
We split training set of TVQA into training and validation set, and serve official validation set as test set in our experiments since test labels are not publicly available.

\subsection{Experimental Details}
\subsubsection{Metrics.}
We follow official objective metrics for AVSD benchmark, including BLEU \cite{papineni-etal-2002-bleu}, METEOR \cite{banerjee-lavie-2005-meteor}, ROUGE-L \cite{lin-2004-rouge}, and CIDEr \cite{Vedantam_2015_CVPR}.
The metrics are formulated to compute the word overlapping between each generated response and reference responses.
\subsubsection{Model Hyperparameters.}
%
The dimension of hidden layer is set to $d = 512$, the number of attention heads for GAT and decoder is set to $K = 8$. Criterions for edge $\mathcal{E}_{st}$ are set to $\tau_s = 0.4$, $\tau_t = 0.2$ for sparse local connection. For GN-GAT, we set distance $n = 1, 2, 3, 4$, and $1, 1, 2, 4$ heads are assigned to each distance, respectively. 
All the hyperparameters were tuned via grid-search over validation set.
%
\subsubsection{Training Details.}
Our model is trained on NVIDIA TITAN V (12GB of memory) GPU with Adam optimizer with $\beta_1 = 0.9, \beta_2 = 0.98$, and $\epsilon = 10^{-9}$.
We adopt a learning rate strategy similar to \cite{Vaswani_2017_NIPS}, and set the learning rate warm-up strategy to $10,000$ training steps and trained model up to $20$ epochs.
We select the batch size of $32$ and dropout rate of $0.3$.
For all experiments, we select the best model that achieves the lowest perplexity on the validation set.
%
During inference, we adopt a beam search with a beam size of $5$ and a length penalty of $1.0$.
The maximum length of output tokens are set to $30$.
The entire framework is implemented with PyTorch.

\begin{table*}[t]
	\centering
	\begin{tabular}{l||c c c c c c c}
		\Xhline{3\arrayrulewidth}
		\multicolumn{8}{c}{AVSD@DSTC7}\\ \Xhline{3\arrayrulewidth}
		Methods                              & BLEU1     & BLEU2     & BLEU3     & BLEU4     & METEOR   & ROUGE-L   & CIDEr     \\ 
		\Xhline{2\arrayrulewidth}
		Baseline \cite{Hori_2019_ICASSP}     & 0.621     & 0.480     & 0.379     & 0.305     & 0.217    & 0.481     & 0.733     \\
		HMA \cite{Le_2019_DSTC7}             & 0.633     & 0.490     & 0.386     & 0.310     & 0.242    & 0.515     & 0.856     \\
		RMFF \cite{Yeh_2019_DSTC7}           & 0.636     & 0.510     & 0.417     & 0.345     & 0.224    & 0.505     & 0.877     \\
		EE-DMN \cite{Lin_2019_DSTC7}         & 0.641     & 0.493     & 0.388     & 0.310     & 0.241    & 0.527     & 0.912     \\
		JMAN \cite{Chu_2020_DSTC8}           & 0.667     & 0.521     & 0.413     & 0.334     & 0.239    & 0.533     & 0.941     \\
		FA-HRED \cite{Nguyen_2019_DSTC7}     & 0.695     & 0.553     & 0.444     & 0.360     & 0.249    & 0.544     & 0.997     \\
		CMU \cite{Sanabria_2019_DSTC7}       & 0.718     & 0.584     & 0.478     & 0.394     & 0.267    & 0.563     & 1.094     \\
		MSTN \cite{Lee_2020_DSTC8}           & -         & -         & -         & 0.377     & 0.275    & 0.566     & 1.115     \\
		JSTL \cite{Hori_2019_Interspeech}    & 0.727     & 0.593     & 0.488     & 0.405     & 0.273    & 0.566     & 1.118     \\
		MTN \cite{Le_2019_ACL}               & 0.731     & 0.597     & 0.490     & 0.406     & 0.271    & 0.564     & 1.127     \\
		MTN-P \cite{Le_2020_DSTC8}           & \bf{0.750}& 0.619     & 0.514     & 0.427     & 0.280    & \bf{0.580}& 1.189     \\ \hline

		SCGA w/o caption                     & 0.702     & 0.588     & 0.481     & 0.398     & 0.265    & 0.546     & 1.059     \\
		SCGA                                 & 0.745     & \bf{0.622}& \bf{0.517}& \bf{0.430}&\bf{0.285}& 0.578     & \bf{1.201}\\	
		\Xhline{3\arrayrulewidth} 
		\multicolumn{8}{c}{AVSD@DSTC8}\\ \Xhline{2\arrayrulewidth}
		MDMN \cite{Xie_2020_DSTC8}           & -       & -       & -       & 0.296   & 0.214  & 0.496   & 0.761 \\
		JMAN \cite{Chu_2020_DSTC8}           & 0.645   & 0.504   & 0.402   & 0.324   & 0.232  & 0.521   & 0.875 \\
		STSGR \cite{Geng_2020_DSTC8}         & -       & -       & -       & 0.357   & 0.267  & 0.553   & 1.004 \\
		MSTN \cite{Lee_2020_DSTC8}           & -       & -       & -       & 0.385   & 0.270  & 0.564   & 1.073 \\
		MTN-P \cite{Le_2020_DSTC8}           & 0.701   & 0.587   & 0.494   & \bf{0.419}   & 0.263  & 0.564   & 1.097 \\ \hline

		SCGA w/o caption                     & 0.675   & 0.559   & 0.459   & 0.377   & 0.269  & 0.555   & 1.024  \\
		SCGA                                 & \bf{0.711}   & \bf{0.593}   & \bf{0.497} & 0.416   & \bf{0.276}  & \bf{0.566}   & \bf{1.123} \\
		\Xhline{3\arrayrulewidth}
	\end{tabular}
	\caption{Experimental results on the test split of AVSD benchmark at DSTC7 and DSTC8 challenges.}
	\label{tab:avsd}
\end{table*}

\subsection{Results on AVSD Benchmark}
Table \ref{tab:avsd} summarizes the experimental results on AVSD dataset. 
We compare SCGA with several baseline methods (please refer to Related Work for description on baseline methods).
For fair comparison, we report the performances of official six reference evaluation on AVSD@DSTC7 and AVSD@DSTC8, without using external data to pretrain the model.
SCGA achieves the state-of-the-art performance against all baseline methods on majority of metrics.
The result indicates that resolving co-reference amongst multiple modalities and capturing fine-grained local-to-global dynamics of video can help to generate quality response to boost model performance.
We also provide experimental results without using caption, which simulates real-word video-grounded dialogue situation; we are only given video context and dialogue history.
Competitive performance of SCGA w/o caption indicates that SCGA is able to reason on contextual cues from video.

\subsection{Ablation Study}
We experiment with several variants of SCGA in order to measure the effectiveness of the proposed key components.
The  first block of Table \ref{tab:ablation} provides the ablation results of structured co-reference resolver. 
While structured co-reference resolver boosts performance significantly, we can see that textual co-reference resolver is more important to integrate informative semantics from key dialogue history, improving CIDEr from $1.161$ to $1.201$.
Without textual co-reference resolver, visual co-reference resolver also cannot work properly since co-reference in question tokens are not resolved.
The second block of Table \ref{tab:ablation} provides the ablation results on spatio-temporal video reasoner.
Without this module, spatial and temporal dynamics of video are implicitly learned through transformer decoder, which shows performance drop of 0.034 in CIDEr.
We further provide results on different distance $n$ for GN-GAT.

\begin{table}[t]
	\centering
	\begin{tabular}{l|| c}
		\Xhline{3\arrayrulewidth}
		 Model Variants                           & CIDEr \\ 
		\Xhline{2\arrayrulewidth}
		Full SCGA                                & 1.201 \\ \hline
		\ \ w/o Textual Co-ref Resolver           & 1.161 \\
		 \ \  w/o Visual Co-ref Resolver        & 1.189 \\
		 \ \ w/o Structured Co-ref Resolver        & 1.152 \\ \hline
		\ \ w/o ST Video Reasoner                 & 1.167 \\
		 \ \ w/ distance $n = 1$                & 1.182 \\
		 \ \ w/ distance $n = [1,2]$             	  & 1.194 \\
		 \ \ w/ distance $n = [1,6]$                  & 1.197 \\
		 \ \ w/ distance $n = [1,8]$                 & 1.187 \\ \hline
		\Xhline{3\arrayrulewidth}
	\end{tabular}
	\caption{Ablation study on model variants of SCGA on the test split of AVSD@DSTC7 benchmark.}
	\label{tab:ablation}
\end{table}

\begin{figure*}[t]
\centering
\includegraphics[width=\textwidth]{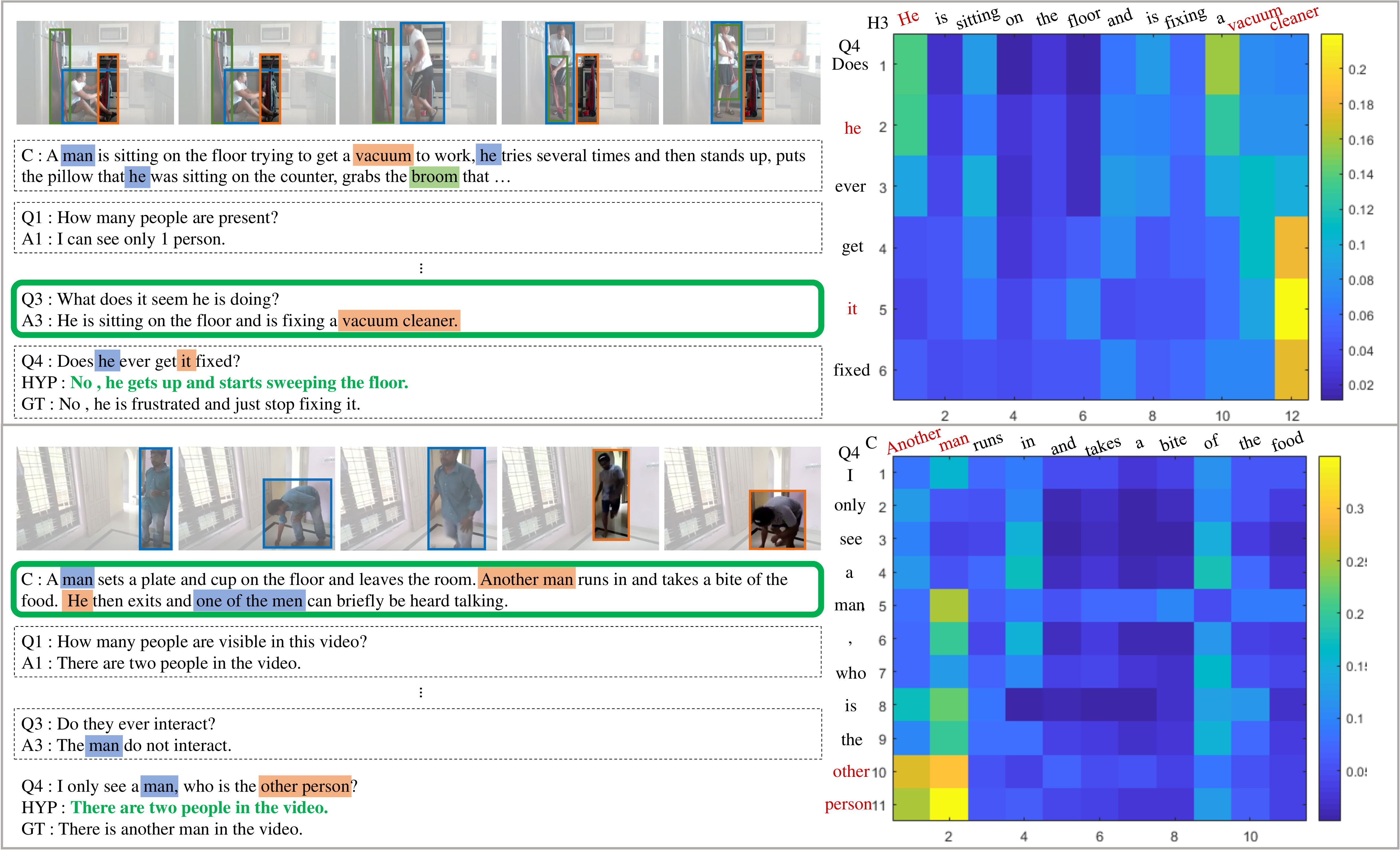}
\caption{Visualization of structured co-reference graph attention (SCGA) in the test split of the AVSD@DSTC7 benchmark.}
\label{fig:qual}
\end{figure*}

\subsection{Results on TVQA Benchmark}
Other than AVSD results, we also report results to TVQA benchmark on our modified setting.
We consider the subtitle corresponding to each video as dialogue history in our experiments on TVQA. 
We compare SCGA with two baseline methods that were reproduced using public codebase.
Table \ref{tab:tvqa} shows SCGA outperforms baseline methods on CIDEr metric.
Subtitles plays an important role in providing clue for answering the question.
Previous approaches to TVQA in multiple-choice setting \cite{Kim_2019_IJCNN,Kim_2019_CVPR,Kim_2020_CVPR} attempted to locate key sentences through temporal attention or temporal localization.
Our results on TVQA demonstrates that SCGA is able to not only resolve co-reference, but also locates a key sentence from subtitle.

\begin{table}[t]
	\centering
	\begin{tabular}{l|| c}
		\Xhline{3\arrayrulewidth}
		Methods                                & CIDEr \\ 
		\Xhline{2\arrayrulewidth}
		Baseline \cite{Hori_2019_ICASSP}          & 0.781     \\
		MTN \cite{Le_2019_ACL}                  & 0.973     \\ \hline
		SCGA                                 & 1.062     \\
		\Xhline{3\arrayrulewidth}
	\end{tabular}
	\caption{Experimental results on the TVQA benchmark.}
	\label{tab:tvqa}
\end{table}

\subsection{Qualitative Analysis}
Figure \ref{fig:qual} visualizes the intermediate functionality of SCGA with samples from test split of the AVSD@DSTC7 benchmark. 
Each example is provided with a selected dialogue history (indicated by green box), learned attention weights for textual and visual co-reference graph.
In figure \ref{fig:qual}, the question of upper example: `Does he ever get it fixed' has a semantic relevance to dialogue history H3: `He is sitting on the floor and fixing a vacuum cleaner' in the green box.
The attention map on the right shows similarity between tokens of Q4 and tokens of H3.
Specifically, `it' token in Q4 refers to `vacuum cleaner' tokens in H3,  showing high attention values.
Through the Visual Co-reference Resolver, the co-referenced objects in video are highlighted from other detected objects where the textual co-reference resolved question makes it easier to find the vacuum cleaner in the video.
The figure below also shows that the high relevant sentence is selected and textual co-reference tokens are enhanced.

\section{Conclusion}
In this paper, VGDS referred to as Structured Co-reference Graph Attention (SCGA) is presented to consider two major challenging issues: (1) How to deduce co-reference among multiple modalities; (2) How to reason on the rich underlying semantic structure of video with complex spatial and temporal dynamics.
SCGA is based on (1) Structured Co-reference Resolver that performs dereferencing via building a structured graph over multiple modalities, (2) Spatio-temporal Video Reasoner that captures both global and local dynamics of video via segmented self-attention layer.
Furthermore, SCGA makes use of pointer network to dynamically replicate parts of the question for decoding the answer sequence.
Our empirical results on AVSD@DSTC7, AVSD@DSTC8 and TVQA benchmarks show that SCGA achieves state-of-the-art performance.

\section{Acknowledgments}

This work was partly supported by Institute for Information \& communications Technology Planning \& Evaluation(IITP) grant funded by the Korea government(MSIT) (2017-0-01780) and partly supported by the MSIT(Ministry of Science, ICT), Korea, under the High-Potential Individuals Global Training Problem(2020-0-01649) supervised by the IITP(Institute for Information \& Communications Technology Planning \& Evaluation) and it has been supported by Microsoft Research.

%

\bibliographystyle{aaai}
\bibliography{bibliography}

\end{document}